# Comparison of different automatic solutions for resection cavity segmentation in postoperative MRI volumes including longitudinal acquisitions

Luca Canalini[1], Jan Klein[1], Nuno Pedrosa de Barros[2], Diana Maria Sima[2], Dorothea Miller[3], Horst Hahn[1]

[1]Fraunhofer MEVIS: Institute for Digital Medicine, Bremen, Germany
[2]icometrix, Leuven, Belgium
[3]Universitätsklinikum Knappschaftskrankenhaus Bochum, Bochum, Germany

**ABSTRACT**

Glioblastoma multiforme (GBM) represents the most common primary brain tumor. After its neurosurgical resection, radiation therapy (RT) is indicated as a postoperative adjuvant treatment to limit a possible regrowth. In RT planning, the target volume is carefully decided including also the resection cavity originated from neurosurgery. The treatment is split into consecutive sessions to reduce side effects. Throughout RT the pathological tissues often change their size and shape, and the target volume defined in the initial planning has to be updated. Magnetic resonance imaging (MRI) data can be acquired before each session to manually redefine the resection cavity contours. However, the manual segmentation of this structure is a tedious and time-consuming task. In this work, we compare five deep learning solutions to automatically segment the resection cavity in postoperative MRI. The proposed methods are based on the same 3D U- Net, widely used to tackle segmentation tasks. We use a dataset of postoperative MRI volumes including also longitudinal acquisitions. Each case counts four MRI sequences and comprises the ground truth of the corresponding resection cavity. Four solutions are trained each with a different MRI sequence. Besides, a method designed with all the available sequences is also presented. In this work, we compare all the solutions to find which one obtains the best DICE index computed between the automatically generated masks and ground truth. Our initial experiments show that the method trained only with the T1 weighted contrast-enhanced MRI sequence achieves the best results, with a median DICE index of 0.81.

## 1. DESCRIPTION OF PURPOSE

Glioblastoma multiforme (GBM) is the most common type of intracranial tumor and has a very poor prognosis [1]. Tumor resection is indicated as the first treatment [2]. However, given the infiltrating nature of GBM, a maximal excision is often unachievable. Therefore, radiotherapy (RT) is normally performed as a postoperative treatment [3]. It uses high energy radiations on a target volume specified in RT planning, to kill pathological cells and to limit tumor recurrence. However, radiation exposure can also damage healthy tissues. To limit side effects, the treatment is usually split into several sessions in which a small dose of the radiation is delivered [1]. The healthy tissue can recover faster from a small fraction of the dose, reducing RT side effects. Besides, careful planning of the target volume is important to limit the radiation dose only to a precise structure.

In intracranial RT planning, the target volume also includes the resection cavity's contours [3]. However, throughout the RT, the anatomical volumes often change. This is also true for the resection cavity, which can go under severe modifications regarding its shape, size, and intensity [4] [5]. An example of the possible alterations is visible in Figure 1, showing two subsequent acquisitions for the same patient. The second and third columns are respectively related to the FLAIR and T1 MRI sequences and show how the intensities of a resection cavity can differ in two consecutive acquisitions. Since it is important to focus the radiation therapy only to the structures of interest, an update of the target volume becomes necessary. Besides computed tomography (CT) scan, postoperative MRI is a valid alternative to obtain updated images of the resection cavity. By observing this data, the contours of the cavity can be manually modified, and the target volume updated. However, the manual segmentation is a tedious work for the physicians [6], who would benefit from having an automatic method to accomplish this task. Despite the importance of the resection cavity in the post-surgical phases, very few solutions have been proposed to automatically delineate its contours. The authors in [6] introduced an automatic method to automatically segment the resection cavity in postoperative MRI. They demonstrated that a convolutional neural network (CNN) can be a valid alternative to the manual segmentation. They used multi-sequence MRI volumes to train their solution. However, they didn't use postoperative MRI volumes acquired at different sessions of the same radiotherapy treatment. An automatic solution tackling the segmentation of the resection cavity also in longitudinal MRI acquisitions is still missing.

In this work, we investigate five different automatic solutions to segment resection cavity in postoperative MRI volumes which also include longitudinal studies. Every method is based on the same CNN based on the 3D U-Net [7], which has become a standard approach in automatic segmentation tasks in medical imaging. Four of the proposed solutions are trained with a distinct MRI sequence, whereas only an approach uses all the available MRI sequences together. A comparison between all the trained models is performed to check which one obtains the best results. This could give a better understanding of which MRI sequences may be the most informative to segment the resection cavity in postoperative MRI.

## 2. MATERIAL AND METHODS

### 2.1 Data

In this work, we utilized the data coming from BraTS 2015 [8], a public dataset including a mixture of pre and postoperative MRI images. The volumes come already skull-stripped, co-registered to the same anatomical template, and interpolated to 1mm³ voxel resolution. For our experiments, we selected 47 postoperative volumes, in which the resection cavity is clearly visible. This data is related to 14 different patients in which high-grade gliomas have been resected. The utilized dataset contains also longitudinal studies for eight patients who have been scanned subsequent times. Each case includes four MRI sequences: T2 weighted (T2), T2 weighted fluid-attenuated inversion recovery (FLAIR), T1 weighted without contrast (T1), T1 weighted gadolinium-enhanced (T1c). An example of the volumes used for our work is visible in Figure 1, in which we can observe the same patient scanned in two subsequent MRI acquisitions. The original challenge focused on the segmentation of other tumor tissues but not on the resection cavity, which therefore wasn't originally segmented. Thus, we manually annotated the structures of interest by looking at the resection cavity on the different MRI sequences. Figures 3 and 4 show two examples of the ground truth annotated for this work (highlighted with green contours).

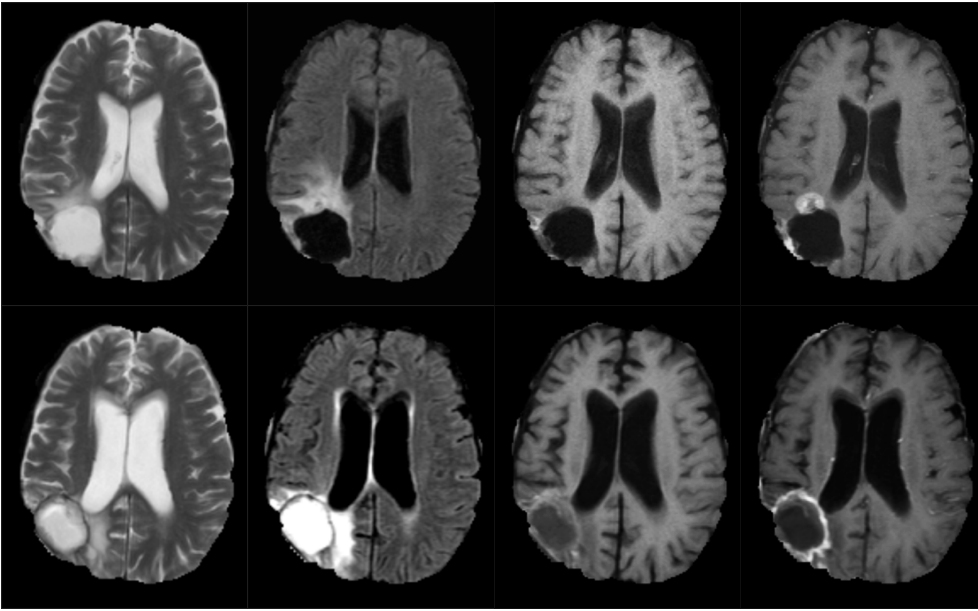

*Figure 1*: Resection cavity changes in two subsequent MRI acquisitions of the same patient. The first row contains four images, each showing different MRI sequences (T2 weighted, FLAIR, T1 weighted and T1 contrast-enhanced) acquired during the same session. The second row shows the same sequences acquired in a subsequent session.

### 2.2 Methods

We here propose five different solutions based on the same CNN architecture. We utilized a 3-leveled 3D U-Net [7] with a receptive field size of 44 by 44 by 44 voxels. The patches containing background are a larger quantity than the ones including resection cavities labels. Thus, to speed the training procedure, the patches composition has been modified: 20% of them include only background voxels, whereas the 80% contain at least a foreground voxel. Moreover, as loss function we used the Tversky Loss to weigh recall higher than precision ($\alpha = 0.2$, $\beta = 0.8$) [9] . It is used to give more penalty on the false negative predictions so that a larger amount of voxels will be segmented as foreground. Even if the false positive predictions may increase, a lower number of the foreground voxels will be missed. All the proposed solutions are trained with a batch of size 5 and the best model is saved every 100 iterations based on the Jaccard coefficient computed on the validation data. Four of the proposed solutions are trained each with a different MRI sequence, and only one has been trained with all the 4 sequences together.

Besides, we utilized a five cross-validation procedure to test our algorithms. The 47 cases are split into five disjoint groups, three of them composed of nine volumes each, two with ten volumes. Each of these disjoint groups represents a test set and the remaining volumes are used to train and validate our solutions. Thus, each proposed method is trained five times and tested always on a disjoint test set. Furthermore, in the inference process, we apply connected component analysis to keep only the largest segmented mask. The Tversky loss function could lead to an over-segmentation of the structure of interest. However, only one resection cavity is present per volume, so we discard smaller structures that may be wrongly segmented.

## 3. RESULTS

We compute the DICE indices between the ground truth and the automatically generated masks [10]. The results of the different approaches are visible in Figure 2, where five box plots show the median DICE values. As we can observe, the model trained with only the T1c MRI sequence obtained a median DICE of 0.81 in our experiments. On the contrary, the 3D U-net trained only on the FLAIR sequence achieved the lowest score median DICE of 0.44. The method using all the four modalities together reaches a median DICE index of 0.79, which is slightly lower than the approach trained with the only T1c, but higher than the other solutions (0.73 and 0.77 are the median DICE values for the solutions trained with T1 and T2). Besides, the Wilcoxon sign-rank test is performed between the outperforming method and each of the other solutions is to verify if the differences are statistically significant. The results of this test are available in Figure 2. The test shows that there are statistically significant differences with a p-value<0.01 between the best solution and methods trained only with T1, FLAIR, and T2. Besides, there is also a statistically significant difference with a p-value<0.05 between the outperforming solution and the model trained with all the sequences. The visual results for the solutions trained with a single MRI sequence are available in Figure 3. Thus, T1c seems to be the most informative sequence to train the proposed method. This may be due to the fact that on T1 contrast-enhanced images the resection cavities usually have uniform characteristics among different acquisitions. In fact, in T1c images, the cerebral spinal fluid (CSF) within the resection cavity often appears hypointense and delimited by a hyperintense border highlighted by the contrast enhancement. Thus, it may be easier for the solution trained with the only T1c to identify unique features to segment the structures of interest. On the FLAIR sequence, the cavities can assume instead very different characteristics even in longitudinal sequences, mainly due to tumor recurrences or radiotherapy treatment [5]. As an example, the second column in Figure 1 shows the same resection cavity observed on the FLAIR sequence during two subsequent acquisitions: The CSF within the resection cavity has very different intensities between the two images. Thus, the method trained only with the FLAIR sequence may not be able to find a specific set of features to correctly characterize the resection cavity. The variability in the intensity could also explain the segmentation errors observed in the second column of Figure 3. The segmented structure (highlighted in orange) includes both hypo and hyperintense parts, mainly because the resection cavity can be characterized by both of them.

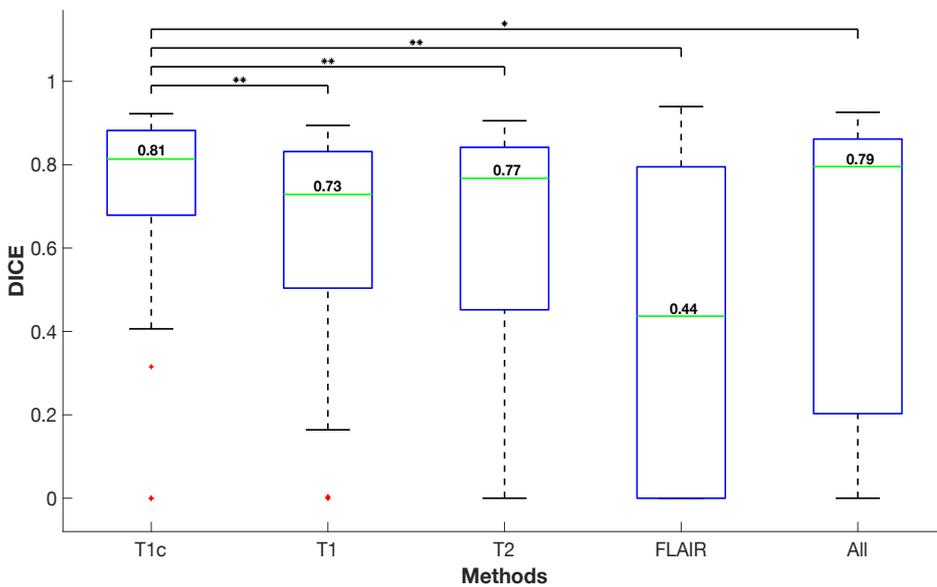

*Figure 2*: DICE results for the five solutions. The methods are listed on the x-axis, whereas the y-axis reports the DICE values. The results of every solution are summarized in the corresponding box plot. The median DICE indices obtained on the whole dataset are indicated by a green line (with the corresponding values written on top of it), whereas the outliers are highlighted with plus-symbols in red color. The model trained only with the T1c sequence is the outperforming solution and shows a statistically significant difference in comparison to the other methods (according to the *Wilcoxon sign rank test*). The lines on top of the graph relate the T1c method with the other solutions. The asterisk on top of each line indicates if the statistically significant difference between the solution trained with T1c and another method is verified with a p-value < 0.05 (one asterisk) or < 0.01 (two asterisks).

Besides, the multisequence approach performs worse than the solution designed with only the T1c sequence. This is surprising because the method trained with multiple MRI sequences should learn more features, potentially leading to a better outcome. However, there may be some MRI sequences that negatively affect the task, especially considering what already discussed for the FLAIR MRI sequence. The visual results for this solution are available in Figure 4.

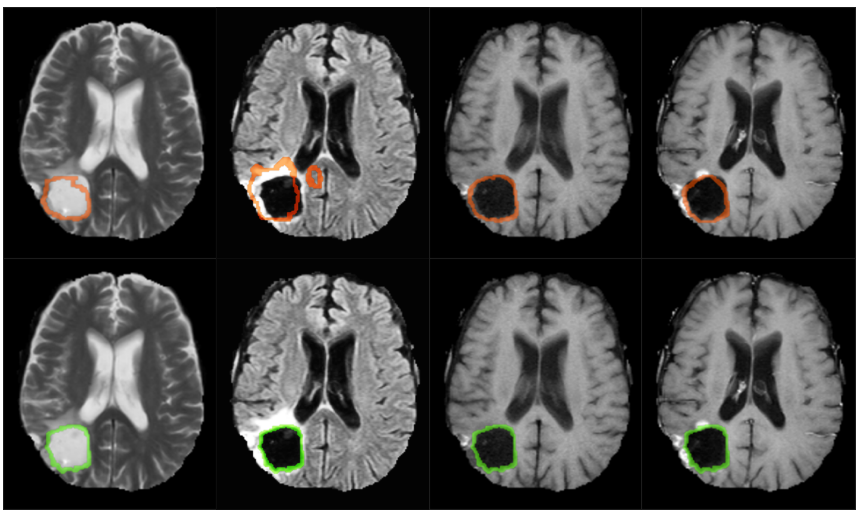

*Figure 3*: Resection cavity segmentation results for the four solutions trained on a different MRI sequence. From left to right, the first row shows the visual segmentation results obtained by the methods trained on T2, FLAIR, T1, and T1c sequence. The second row displays the corresponding ground truth observed in the four different MRI sequences. The automatically generated segmentation of the resection cavity is highlighted by an orange border (first row). The ground truth is highlighted in green in the second row.

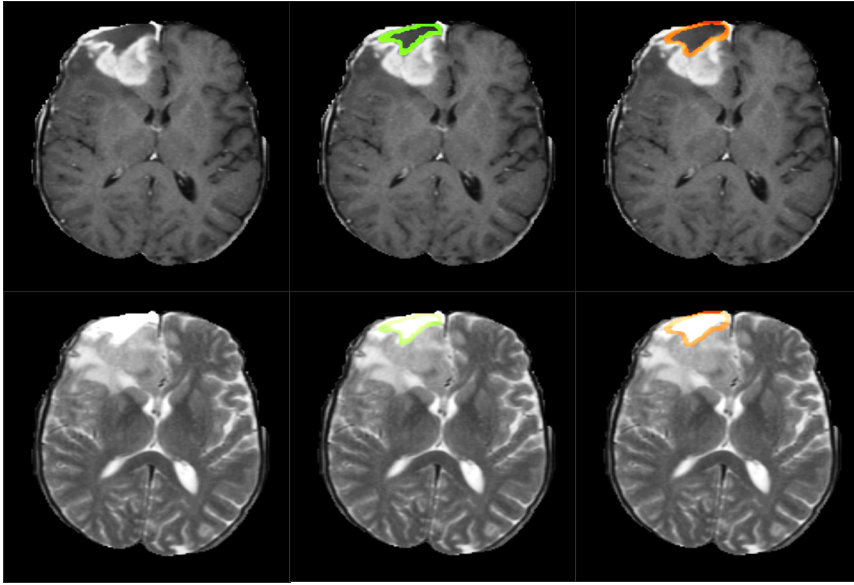

*Figure 4*: Example of resection cavity segmentation for the solution trained on multisequence MRI. The results are shown on the T1c and T2 sequences (respectively first and second row). The first column displays the original volume, the second one the ground truth (in green) and the last one shows the automatically generated mask (in orange).

## 4. NEW OR BREAKTHROUGH WORK TO BE PRESENTED

To the best of our knowledge, we proposed the first comparison between different solutions aimed to segment the resection cavity in postoperative MRI including also longitudinal studies. We compare five approaches based on the same 3D U-Net architecture, each trained on five different MRI sequences combination. Four methods are designed only with a single MRI sequence, whereas one approach is trained with all the four different sequences. Our experiments show that training performed with only T1 post-contrast MRI sequence achieves the best results, also when compared with the multi-sequence approach. Besides, the outperforming method shows a statistically significant difference compared to the other methods. On the contrary, the solution trained only with the FLAIR sequence achieves the lowest DICE score.
***This work is not being, or has not been, submitted for publication or presentation elsewhere.***

## 5. CONCLUSIONS

We have proposed a comparison between five different solutions based on the same 3D U-Net architecture to segment the resection cavity in postoperative MRI volumes including longitudinal studies. Each method is trained with a different combination of MRI sequences. Our experiments showed that the solution trained with only the T1c obtains the best results. This represents a good starting point for further investigations about resection cavity segmentation, where very few automatic solutions have been proposed so far. Instead of using all the available MRI sequences, the next solutions may utilize only the most informative ones. For future work, different deep learning architectures could be tested to verify which one would be the best to segment the desired output. Besides, multi-label approaches could be investigated, in which other pathological tissues (for instance edema and active tumor) are segmented together with the resection cavity.